\documentclass{article}

\usepackage{microtype}
\usepackage{graphicx}
\usepackage{subfigure}
\usepackage{booktabs} % for professional tables
\usepackage{xcolor}
\usepackage{amsfonts}
\usepackage{amsmath}
\usepackage{float}

\usepackage{hyperref}

\usepackage[accepted]{icml2020}

\icmltitlerunning{Optimal Transport using GANs for Lineage Tracing}

\begin{document}

\twocolumn[
\icmltitle{Optimal Transport using GANs for Lineage Tracing}



\icmlsetsymbol{equal}{*}

\begin{icmlauthorlist}
\icmlauthor{Neha Prasad}{to,equal}
\icmlauthor{Karren D.  Yang}{to,equal}
\icmlauthor{Caroline Uhler}{to}
\end{icmlauthorlist}

\icmlaffiliation{to}{Laboratory for Information and Decision Systems, Massachusetts Institute of Technology, Cambridge, MA, USA}

\icmlcorrespondingauthor{Caroline Uhler}{cuhler@mit.edu}

\icmlkeywords{Computational biology, lineage tracing, optimal transport, GANs}

\vskip 0.3in
]
\printAffiliationsAndNotice{\icmlEqualContribution} % otherwise use the standard text.

\begin{abstract} 
  In this paper, we present Super-OT, a novel approach to computational lineage tracing that combines a supervised learning framework with optimal transport based on Generative Adversarial Networks (GANs). Unlike previous approaches to lineage tracing, Super-OT has the flexibility to integrate paired data. We benchmark Super-OT based on single-cell RNA-seq data against Waddington-OT, a popular approach for lineage tracing that also employs optimal transport. We show that Super-OT achieves gains over Waddington-OT in predicting the class outcome of cells during differentiation, since it allows the integration of additional information during \mbox{training.} 
  
\end{abstract}

\section{Introduction}

A major goal of developmental biology is understanding the gene programs that drive the differentation of progenitor cells into mature cells. Although single-cell RNA sequencing technologies enable collecting large quantities of single-cell gene expression data, these experiments tend to be destructive to samples: most experiments only take a snapshot of a single cell lineage at one time point. This makes it challenging to trace the lineage of individual cells and necessitates the development of novel computational and experimental strategies to follow individual cell trajectories backwards and forwards in time.

In recent years, various computational methods have been developed to trace pseudo-lineages of cells given single-cell gene expression data \cite{monacle, monacle2, slice, waterfall, tscan, scuba, wanderlust, wishbone, paga, weinreb2018fundamental, schiebinger2017reconstruction}. All these methods are unsupervised and depend heavily on the assumption that cells close in gene expression space are more likely to belong to the same lineage, which is often not sufficient for accurately predicting cell fate decisions. For example, \citet{Weinreb467886} recently developed an experimental assay to track families of cells (i.e., cell clones) during hematopoiesis. Their work revealed a considerable gap between the pseudo-lineages that are reconstructed computationally and real cell trajectories. In light of recent experimental methods for providing partial information about single-cell lineages, there is a need for computational methods to integrate these types of data with existing unsupervised lineage tracing frameworks.

In this work, we develop to our knowledge the first model for computational lineage tracing that can integrate additional information such as the clonal data of \citet{Weinreb467886}. Since their data does not provide full information about single cell lineages (e.g., due to multiple cells from the same clone that cannot be disambiguated), our work combines a supervised learning framework with the principles of optimal transport to predict individual cell lineages. Optimal transport has emerged as a powerful method for learning couplings between cells and has many advantages in computational lineage tracing compared to other methods \cite{schiebinger2017reconstruction}. The best-known computational approach for lineage tracing based on optimal transport, Waddington-OT, uses an iterative scaling algorithm and cannot be adapted easily to integrate additional labeled data \cite{schiebinger2017reconstruction}. Recent approaches for performing optimal transport based on GANs have also emerged \cite{yang2018scalable}, but these have not been rigorously compared against Waddington-OT and were also not adapted to integrate additional labeled data.

\textbf{Contribution.} We propose a new framework for computational lineage tracing, which we name \emph{Super-OT}, that combines a \emph{super}vised learning framework with optimal transport (\emph{OT}) based on GANs. We apply this framework to perform lineage tracing on the dataset of  \citet{Weinreb467886}. While Waddington-OT marginally outperforms GAN-based optimal transport in the completely unsupervised setting, we find that Super-OT outperforms Waddington-OT in predicting the class outcome of cells during differentiation by integrating additional information during training. We conclude that GAN-based optimal transport is a practical approach for integrating novel types of experimental data in lineage tracing with a principled computational approach.

\section{Method}

\subsection{Problem Formulation.}

Let $P_t$ and $P_{t+1}$ denote cell distributions over the gene expression space $\mathbb{R}^d$ at times $t$ and $t+1$, respectively. We formulate computational lineage tracing as the problem of learning a transport map $T: \mathbb{R}^d \rightarrow \mathbb{R}^d$ that pushes $P_t$ to $P_{t+1}$, i.e., if $X_t \sim P_t$ and $X_{t+1} \sim P_{t+1}$, then $T(X_t) \sim P_{t+1}$. We assume that we are given samples from $P_t$, $P_{t+1}$ and their joint distribution $P_{t, t+1}$. We denote the observed empirical distributions by $\hat{P}_t$, $\hat{P}_{t+1}$ and $\hat{P}_{t, t+1}$. 

\subsection{Optimal Transport using GANs.} 

There are many possible maps satisfying $T(X_t) \sim P_{t+1}$. One way to constrain the solution space is to solve the Monge optimal transport problem: find the map that minimizes some transportation cost $c: \mathbb{R}^d \times \mathbb{R}^d \rightarrow \mathbb{R}^+$, i.e., 
\begin{equation}\label{eq:Monge}
    \min_T\quad \mathbb{E}_{X_t \sim P_t} c(X_{t+1}, T(X_t))
\end{equation}
subject to $T(X_t) \sim P_{t+1}$. In the context of lineage tracing, $c$ is often chosen to be the Euclidean distance $c(x, y) = ||x-y||_2$, which encourages the map to match cells that are closer together in the gene expression space. The intuition is that cells closer in gene expression space should be more likely to belong to the same lineage. Alternatively, the cost can also be computed in a feature space learned by an autoencoder, as proposed by \citet{Yang455469}.

In practice, Equation (\ref{eq:Monge}) is challenging to optimize due to the constraint $T(X_t) \sim P_{t+1}$. We follow \citet{yang2018scalable} and relax the hard constraint using a divergence that can be optimized using an adversarial approach. Specifically, we instead consider solving
\begin{equation}\label{eq:GAN-OT}
    \min_T \quad \lambda_{1} \mathcal{L}_{TRANS} + \mathcal{L}_{GAN},
\end{equation}
where
%\begin{equation*}
\begin{align*}
\mathcal{L}_{TRANS} &= \mathbb{E}_{X_t \sim P_t} c(X_{t+1}, T(X_t)), \quad\textrm{and}\\
\mathcal{L}_{GAN} &= \max_{D: ~\mathbb{R}^d \rightarrow (0, 1)}~ \mathbb{E}_{X_{t+1} \sim P_{t+1}}[\log D(X_{t+1})] \\
&\quad + \mathbb{E}_{X_t \sim P_t}[\log(1 - D(T(X_t)))],
\end{align*}
and $\lambda_{1} > 0$ is a hyperparameter. In practice, we can parameterize $T, D$ using neural networks and minimize the loss with respect to the observed empirical distributions.

\subsection{Super-OT} 

Existing methods for lineage tracing based on optimal transport do not leverage labeled information. Our approach extends the GAN-based optimal transport framework to pair cells between two distinct time points that belong to the same clonal family. We consider an additional loss
\begin{equation}
    \mathcal{L}_{SUPER} = \mathbb{E}_{(X_t, X_{t+1}) \sim \hat{P}_{t, t+1}} ||X_t-X_{t+1}||^2_2,
\end{equation}
which ensures that points belonging to the same clonal family are mapped to each other. 
The final objective then becomes
\begin{equation}\label{eq:objective}
    \min_T \quad \lambda_{1} \mathcal{L}_{TRANS} + \mathcal{L}_{GAN} + \lambda_2 \mathcal{L}_{SUPER},
\end{equation}
where $\lambda_1, \lambda_2 > 0$ are hyperparameters.

\section{Experiments}

\subsection{Dataset}
We use the single-cell gene expression dataset of mouse bone marrow cells from \citet{Weinreb467886}. The dataset contains 130,887 samples across 25,289 genes, sampled from three separate time points (days 2, 4, and 6). We let $P_{t}$ corresponds to undifferentiated day 2 cells, while we let $P_{t + 1}$ correspond to differentiated days 4/6 cells. For our experiments, we consider a subset of the data containing neutrophils, monocytes, and their progenitors. This simplified the dataset to contain 1,527 day 2 cells and 39,401 day 4/6 neutrophils and monocytes. Cell family information is provided in the form of a binary matrix in which rows represent cells and columns represent clones, where an entry of $1$ indicates that a cell belongs to a particular clone. For the supervised labeling of the data, we consider a day 2 cell to be paired with a day 4/6 cell if they belong to the same cell family and the day 4/6 cell is in the majority class (neutrophil versus monocyte) for that family.

%The third and last table is \textbf{clone\_annotation\_in\_vitro.mtx}, which is a binary matrix representing the clonal membership of each cell, with the matrix containing information on 5,864 clones for the 130,887 cell samples. 

\subsection{Model and Training.}

We implemented our model in PyTorch \cite{NEURIPS2019_9015}. The transport $T$ uses four linear layers with hidden dimension of 1000 with ReLU activations. The discriminator $D$ also consists of four linear layers with batch normalization and ReLU activations. The dataset reports the number of transcripts (UMIs) for each gene in each cell, after total counts normalization (i.e., L1 normalization on cells). We add an additional layer of preprocessing by scaling each feature (gene) between the range $\left[0, 1\right]$ with L2 normalization. We also apply PCA \cite{scikit-learn} to reduce the dimensionality of the gene expression vectors to 100. Note that this reduced dimensionality justifies using Euclidean distance for $\mathcal{L}_{TRANS}$. The networks are trained jointly using the loss in Equation \ref{eq:objective} with different numbers of paired samples. For the transport cost, we used $\lambda_{1}=0.6$, because we found that $\lambda_{1} > 0.6$ would cause the model to not converge. Similar to other GANs \cite{gulrajani2017improved}, we also include a gradient penalty on the discriminator to improve training stability.  Parameters are optimized using Adam \cite{kingma2014adam} with learning rates set to 0.0001 for each model. We run the model for a maximum of 100,000 epochs, or until the loss has sufficiently converged. We train each setting with 80\% of the cells, and leave the remaining 20\% for evaluation.

We evaluate the performance of Super-OT with different numbers of labeled pairs of data: 300, 600, and 900. 

\subsection{Benchmarks}

We compare Super-OT against the following benchmarks:
\begin{itemize}
    \item Waddington-OT \cite{schiebinger2017reconstruction}, which we denote by WOT in Table~\ref{table:accuracy}, a lineage tracing method that solves the optimal transport problem using an iterative scaling algorithm.
    \item Conditional GAN \cite{mirza2014conditional}: A model that generates the day 4/6 cells from day 2 conditioned on class (monocyte/neutrophil).
    \item GAN-based Optimal Transport: Conditional GAN combined with transport cost.
    \item Supervised: A regression model that maps each day 2 cell to a corresponding day 4/6 cell in its clonal family. This method uses all paired information available and should be considered an upper bound on performance.
\end{itemize}

\subsection{Evaluation Criteria.} \label{Evaluation}

We evaluate each model on its class prediction accuracy in transporting day 2 cells to day 4/6 monocytes versus neutrophils. Specifically, from the dataset clonal information, we determine whether each day 2 cell is more likely to become a neutrophil or a monocyte and assign this label to the cell. We train a logistic regression model \cite{scikit-learn} to classify between real day 4/6 monocytes and neutrophils, and obtain the predicted class for each transported day 2 cell. We then compare these predicted results with the assigned labels. The accuracy reflects the ability of the models to correctly predict the direction of the trajectory of the day 2 test cells towards their differentiated class.

Waddington-OT differs from the other models in that it produces a probabilistic coupling between cells: each day~2 cell is assigned to a distribution over the day 4/6 cells \cite{schiebinger2017reconstruction}. To obtain a class prediction for each day 2 cell, we assign a label depending on whether the majority of the distribution is assigned to day 4/6 monocytes or neutrophils. For fairer comparison with the other models, we alternatively assign a label depending on whether the majority of the distribution is assigned to predicted monocytes or neutrophils using a logistic regression model. This check ensures that differences in performance are not due to the logistic classifier.

\begin{table}[t]
    \caption{Comparison of prediction accuracy of different models. Higher is better. For each deep learning model, we perform three separate runs over different train/test splits and report the average of the best results on the held-out data.}
    \label{table:accuracy}
    \vskip 0.15in
    \begin{center}
    \begin{small}
    \begin{sc}
    \begin{tabular}{lll}
    \hline
    \textbf{Setting}  & \textbf{Accuracy} \\
    \midrule
    WOT: Predicted Labels & 0.6535 \\
    \hline
    WOT: Real Labels & 0.6531 \\
    \hline
    Conditional GAN & 0.5982\\
    \hline
    GAN-Based OT & 0.6219 \\ 
    \hline
    \hline
    Super-OT: 300 & 0.6731 \\ 
    \hline
    Super-OT: 600 & 0.6856 \\
    \hline
    Super-OT: 900 & \bf{0.7188}  \\
    \hline
    \hline
    Supervised & 0.7534 \\
    \hline
    \bottomrule
    \end{tabular}
\end{sc}
\end{small}
\end{center}
\vskip -0.1in
\end{table}

\subsection{Results}

We report the class prediction accuracy of the different models in Table \ref{table:accuracy}. For Super-OT, we report results with different numbers of paired training points (e.g., 300, 600, 900.) Out of the unsupervised baseline methods, Waddington-OT outperforms the conditional GAN and GAN-based optimal transport. This is expected because Waddington-OT is based on an iterative scaling algorithm with convergence guarantees, while deep models solve non-convex optimization problems and do not have such guarantees. However, GAN-based optimal transport is flexible and can be adapted to handle additional losses, such as our paired data supervised loss. Super-OT, which is our extension of GAN-based optimal transport with some labeled data, outperforms all the baselines including Waddington-OT. Performance increases as more labeled data is added and is upper bounded by the fully supervised model.

\textbf{Ablations.}
Since Super-OT integrates supervised training with optimal transport, we investigate the importance of each component of the loss function.

\begin{itemize}
    \item Supervised Loss ($\mathcal{L}_{SUPER}$): As shown in Table \ref{table:accuracy}, decreasing the number of labeled points decreases the accuracy of the method.
    \item Transport cost ($\mathcal{L}_{TRANS}$): As shown in Table \ref{transport-cost}, removing the transport cost decreases the accuracy of the method for any given number of labeled points.
\end{itemize}

\begin{table}[t]
    \caption{Comparison of prediction accuracy for Super-OT: 300, Super-OT: 600, Super-OT: 900 with and without the transport cost.}
    \label{transport-cost}
    \vskip 0.15in
    \begin{center}
    \begin{small}
    \begin{sc}
    \begin{tabular}{lll}
    \hline
    \textbf{Setting}  & \textbf{Transport Cost} & \textbf{No Transport Cost} \\
    \midrule
    Super-OT: 300 & 0.6731 & 0.6663\\ 
    \hline
    Super-OT: 600 & 0.6856 & 0.6810 \\
    \hline
    Super-OT: 900 & 0.7188 &  0.6997 \\
    \hline
    \bottomrule
    \end{tabular}
\end{sc}
\end{small}
\end{center}
\vskip -0.1in
\end{table}

\begin{figure*}[!t]
\begin{minipage}{.5\textwidth}
  \centering
  \includegraphics[width=\textwidth]{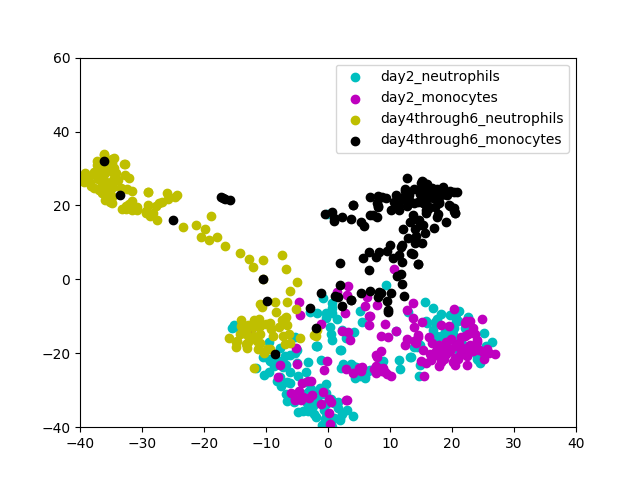}
\end{minipage}%
\begin{minipage}{.5\textwidth}
  \centering
  \includegraphics[width=\textwidth]{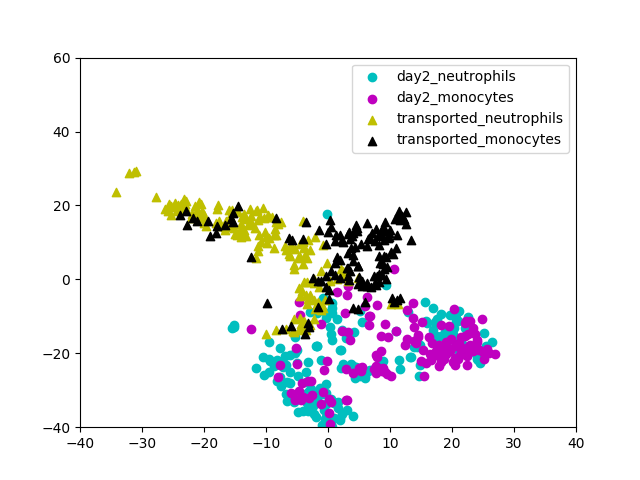}
\end{minipage}%
\caption{Super-OT: 600 with Transport Cost, Actual Neutrophil/Monocyte Distributions (left), Transported Neutrophil/Monocyte Distributions (right). The plot shows how day 2 neutrophils (blue) and monocytes (pink) transform into differentiated neutrophils (yellow) and monocytes (black). }
\label{fig:transport-tSNE}

\end{figure*}

\textbf{Visualization.} We visualize the day 2, day 4/6 and transported test cells for our Super-OT model using t-SNE \cite{scikit-learn} in Figure \ref{fig:transport-tSNE}. We see that the model is separating the Neutrophil/Monocyte distributions correctly, and that there is overlap between the transported cell distribution and the ground truth. However, the transported cell clusters are closer to each other than compared to the real cell clusters. One explanation is that differentiation is not yet deterministic on day 2: despite overlapping in gene expression space, some day 2 cells become monocytes and others become neutrophils. Therefore the model transports cells to a weighted average between the two clusters. In future work, incorporating stochasticity into the transport map may improve separation of the transported clusters.

\subsection{Differential Gene Analysis}
To determine how faithful Super-OT is for modeling the molecular programs that drive differentiation to monocytes or neutrophils, we performed a differential gene expression analysis to determine which genes are predicted to be
the most {differentially expressed} between the 
neutrophil and monocyte distributions. We then compared these gene programs to the actual genes that are differentially expressed between the distributions.
Specifically, we used the networks corresponding to the best accuracy values in Table \ref{table:accuracy}. Note that after preprocessing, all of these cells have dimension 100, and so we used the inverse PCA transform to revert cells to their original 25289-dimensional gene space.
From here, we iterated through each of
the genes and performed a t-test to determine the differentially-expressed genes, setting our p-value threshold at $10^{-6}$.

\iffalse
from scipy's stats package 
between each of 
day4\_6\_neutrophils[i] and day4\_6\_monocytes[i], 
transported\_neutrophils[i] and transported\_monocytes[i], 
where $i$ denotes the $i$th gene $ \forall i \in [0, 25,288]$
(since we have a total of 25, 289 genes). 

To ensure that we
found the genes that were the most differentially 
expressed, we used a $p$-value of $10^{-6}$.
\fi
Finally, we selected the genes that were common among all three runs for our analysis. Table \ref{differentialgenes} lists the precision and recalls for the different settings. We found that there was considerable overlap between the predicted and actual differentially expressed genes. The precision / recall values were robust to different $p$-values. Importantly, the differentially expressed genes found by Super-OT were much more stable than those found using GAN-based OT.

\begin{table}[!tb]
  \caption{Differential Gene Expression Analysis: Comparison between the precision / recall values for GAN-Based OT, Super-OT: 300, Super-OT: 600, and Super-OT: 900}
    \vskip 0.15in
    \begin{center}
    \begin{small}
    \begin{sc}
    \begin{tabular}{llll}
    \hline
    \textbf{Setting}  & \textbf{\# of Genes} & \textbf{Precision} & \textbf{Recall}  \\
    \midrule
    GAN-Based OT & 12 & 1.0 & 0.006726\\
    \midrule
    Super-OT: 300 & 3084 & 0.4212 & 0.6969\\ 
    \hline
    Super-OT: 600 & 5041 & 0.3210 & 0.8662\\
    \hline
    Super-OT: 900 & 3021 & 0.3923 & 0.6096\\
    \hline
\bottomrule
\end{tabular}
    \label{differentialgenes}
\end{sc}
\end{small}
\end{center}
\vskip -0.1in

\end{table}

\section{Discussion}
In this work, we proposed Super-OT, a new computational lineage tracing method that combines a supervised learning framework with GAN-based optimal transport. Our framework can easily be extended to map cells between multiple time points, in which we use the timepoint as extra conditioning information. Super-OT achieves gains over Waddington-OT in predicting the class outcome of cells during differentiation by integrating additional information during training. In future work, it would be interesting to analyze how our framework performs when a scaling factor is incorporated to model growth (replication) or shrinkage (death) of cells as proposed by \citet{yang2018scalable}. In addition, it would be interesting to investigate how one could incorporate stochasticity into Super-OT, instead of using a deterministic map, to achieve better cluster separation. 

\section*{Acknowledgments}
 Karren D.~Yang was supported by a National Science Foundation (NSF) Graduate Research Fellowship and ONR (N00014-18-1-2765 and  N00014-18-1-2765). Caroline Uhler  was  partially  supported  by NSF (DMS-1651995), ONR (N00014-17-1-2147 and N00014-18-1-2765), IBM, MIT J-WAFS and J-Clinic, as well as a Simons Investigator Award.

%\newpage
\bibliography{references.bib}
\bibliographystyle{icml2020}

\end{document}